\documentclass{sig-alternate-05-2015}

\usepackage{graphicx}
\usepackage{rotating}
\usepackage{multirow}
\usepackage{algorithm}
\usepackage{algorithmic}
\usepackage{amsmath,amssymb,amsfonts}
\usepackage{booktabs}
\begin{document}

\setcopyright{acmcopyright}

\doi{10.475/123_4}

\isbn{123-4567-24-567/08/06}

\conferenceinfo{PLDI '13}{June 16--19, 2013, Seattle, WA, USA}

\acmPrice{\$15.00}

%
\conferenceinfo{WOODSTOCK}{'97 El Paso, Texas USA}

\title{A Novel Multi-Task Tensor Correlation Neural Network for Facial Attribute Prediction }

\author{
Mingxing Duan$^{1, 2}$,
Kenli Li$^1$,
Qi Tian$^3$,
\\
$^1$ College of Information and Engineering, Hunan University, Changsha, China \\
$^2$ School of Computer Science, National University of Defense Technology, Changsha, China\\
$^3$ Department of Computer Science, University of Texas at San Antonio, USA  \\
duanmingxing16@nudt.edu.cn,
lkl@hnu.edu.cn,
qi.tian@utsa.edu
}
%

\maketitle
\begin{abstract}
Face multi-attribute prediction benefits substantially from multi-task learning (MTL), which learns multiple face attributes simultaneously to achieve shared or mutually related representations of different attributes. The most widely used MTL convolutional neural network is heuristically or empirically designed by sharing all of the convolutional layers and splitting at the fully connected layers for task-specific losses. However, it is improper to view all low and mid-level features for different attributes as being the same, especially when these attributes are only loosely related. In this paper, we propose a novel multi-attribute tensor correlation neural network (MTCN) for face attribute prediction. The structure shares the information in low-level features ({\em e.g.,} the first two convolutional layers) but splits that in high-level features ({\em e.g.,} from the third convolutional layer to the fully connected layer). At the same time, during high-level feature extraction, each subnetwork ({\em e.g.,} Age-Net, Gender-Net, ..., and Smile-Net) excavates closely related features from other networks to enhance its features. Then, we project the features of the C9 layers of the fine-tuned subnetworks into a highly correlated space by using a novel tensor correlation analysis algorithm (NTCCA). The final face attribute prediction is made based on the correlation matrix. Experimental results on benchmarks with multiple face attributes (CelebA and LFWA) show that the proposed approach has superior performance compared to state-of-the-art methods.
\end{abstract}

\printccsdesc


\section{Introduction}
Human face attribute estimation has received a large amount of attention in visual recognition research because a face attribute provides a wide variety of salient information, such as a person's identity, age, race, gender, hair style, and clothing. Recently, many researchers have used face attributes in real-life applications, such as (i) identification, surveillance, and internet access control \cite{Fu2010Age}, \cite{Liu2015Age}, {\em e.g.,} automatic detection of juveniles on the Internet, or surveillance at unusual hours or in unusual places; (ii) face retrieval \cite{DBLP:journals/corr/RanjanPC16}, \cite{5871641}, {\em e.g.,} automatic finding of person(s) of interest with provided attributes in a database; and (iii) social media \cite{6035718}, \cite{Qi2009Learning}, {\em e.g.,} automatic recommendation of makeup or hair styles.

\begin{figure}[t]
\centering
\setlength{\fboxsep}{0pt}
\setlength{\fboxrule}{0pt}
\includegraphics[height =1.2in,width=2.4in]{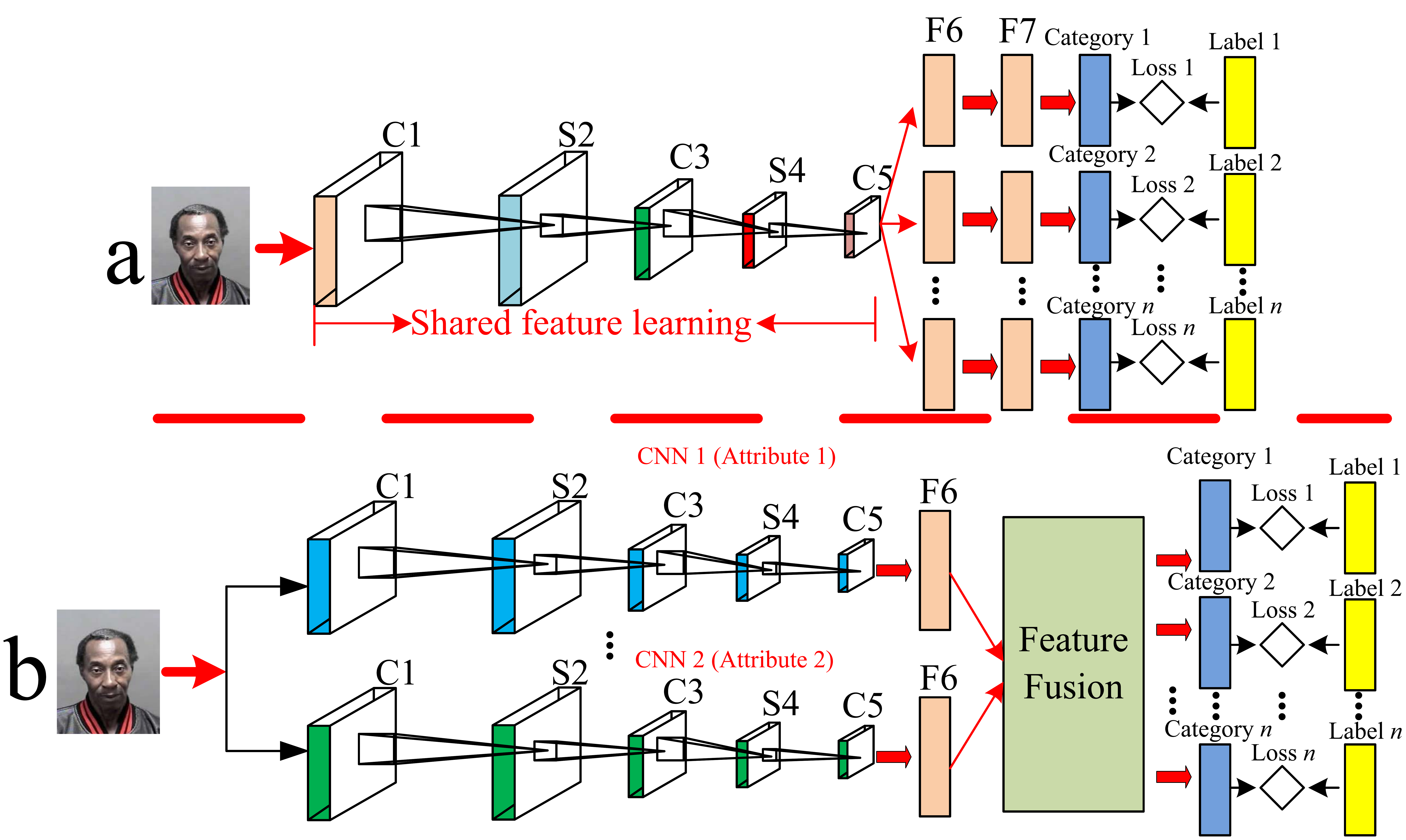} \\
\DeclareGraphicsExtensions.
\caption{The methods used for face attribute prediction. }
\label{fig:1}
\end{figure}

In spite of the recent progress in face attribute estimation \cite{Kwon1994Age}, \cite{8082561}, \cite{Rothe2016Deep}, much of the prior work has been limited to predicting a single face attribute or learning a separate model to estimate each face attribute. As is known, face attributes are strongly related, such as goatee and male, heavy makeup and wearing lipstick and other relationships, and fully exploiting the correlation can help each task be learned better. A joint estimation of face attributes can address this embarrassing situation by exploring attribute correlations \cite{7410782}, \cite{7254184}, \cite{ehrlich2016facial}, \cite{8170321}, \cite{Han2017Heterogeneous}, and can achieve state-of-the-art performance for some face attribute predictions. These methods can be divided into two categories: multi-task learning (MTL) and multi-CNN fusion. Learning tasks in parallel while utilizing shared information to seek correlations is the main point of MTL, and in most work, as in Fig. \ref{fig:1} (a), there is shared representation from the first convolution layer to the last fully connected layer. However, \cite{misra2016cross} proved that it is not sensible to completely share representations between tasks, and these approaches ignore the differences and interactions among these attributes. In contrast, as illustrated in Fig. \ref{fig:1} (b), although the multi-CNN fusion method addresses the differences and explores the correlation through the output of the fully connected layer, it is difficult to realize end-to-end learning and neglect the correlation between the intermediate different attribute features.

Further, \cite{hand2017attributes} proposed a multi-task deep convolutional neural network for attribute prediction via sharing the lower layers in the CNN instead of all of its layers, and this process achieved better performance on many attributes. It follows that attributes and objects share a low-dimensional representation, which allows the object classifier to be regularized\cite{5995543}. This approach does not fully explore the correlation among the high-level features of the face attributes, and each face attribute prediction should not only consider their difference but also utilize attribute correlation. Motivated by the analysis above, we propose a novel multi-task learning structure for face attributes that shares information in the lower-level feature layers and learns the differences and correlations among the high-level features.

At the same time, a large amount of work has proven that each face attribute estimation can be enhanced based on others, such as gender estimation based on smile dynamics \cite{7755833}, age estimation combined with smile dynamics \cite{7060650}, and age estimation affected by gender and race \cite{5543609}. Although some face attribute predictions benefit from others, the degrees of influence on an attribute among other different attributes are not the same, and a unified correlation mechanism might not be appropriate. Consequently, a perfect face attribute should not only adequately seek the differences and correlations among the attributes but should also attempt to exploit the specific degrees of correlation among them. A novel tensor correlation analysis algorithm (NTCCA) is proposed to exploit the detailed correlations of the high-level features from the C9 layer of the fine-tuned subnetworks. A generalization matrix is utilized to ensure that each projected feature space is more highly correlated, which makes each face attribute fully exploit a maximal benefit from the others. Parts of the training dataset are used to train this matrix, and the experimental results indicate that this operation makes the whole system more stable and robust.

In this paper, a multi-task correlation learning neural network (MTCN) is proposed to predict face attributes. The system tries its best to capture the correlations among these attributes, which includes sharing information in low-level feature layers and splitting that in the high-level feature layers while extracting related information from other subnetworks to enhance its useful features and, finally, excavating the correlation among the C9 layers with a novel tensor correlation analysis algorithm (NTCCA). The detailed process of multi-task correlation learning is shown in Fig. \ref{fig:2}. We first train the subnetwork with the corresponding attributes on CelebA or LFWA, and the fine-tuned MTCN is used to predict the attributes of CelebA or LFWA, which is our MTCN without NTCCA. Then, the features of the fine-tuned subnetworks for an image in the C9 layer are built into a tensor, and NTCCA is utilized to project the original features into the highly correlated feature space. Finally, CelebA and LFWA are used to verify the performance of the fine-tuned MTCN. The experimental results demonstrate that our approach significantly outperforms the state-of-the-art methods by achieving average accuracies of 92.97\% and 87.96\% on CelebA and LFWA, respectively.
\section{Related Work }
\subsection{Tensor Canonical Correlation Analysis}
The $n$-mode product of ${\cal X}$ with the matrix $U \in {R^{{J_n} \times {I_n}}}$ is then denoted as ${\cal M} = {\cal X} \times {}_nU$, which is an ${I_1} \times {I_2} \cdot  \cdot  \cdot {I_{n - 1}} \times {J_n} \times {I_{n + 1}} \cdot  \cdot  \cdot  \times {I_N}$ tensor with the element
\begin{equation}\label{}
M({i_1},...,{i_{n - 1}},{j_p},{i_{n + 1}},...,{i_N})= \sum\limits_{{i_n} = 1}^{{I_n}} {{\cal X}({i_1},{i_2},...,{i_N})U({j_n},{i_n})}.
\end{equation}

The product of ${\cal X}$ and a sequence of matrices $\{ {U_n} \in {R^{{J_n} \times {I_n}}}\} _{n = 1}^N$ is a ${J_1} \times {J_2} \times  \cdot  \cdot  \cdot  \times {J_N}$ tensor denoted by
\begin{equation}\label{}
{\cal M} = \;{\cal X} \times {}_1{U_1} \times {}_2{U_2} \cdot  \cdot  \cdot  \times {}_N{U_N}.
\end{equation}


The CANDECOMP / PARAFAC (CP) decomposition decomposes an $N$th-order tensor, ${\cal X} \in {R^{{I_1} \times {I_2} \times  \cdot  \cdot  \cdot  \times {I_N}}}$, into a linear combination of terms, $a_r^{(1)} \circ a_r^{(2)} \circ  \cdot  \cdot  \cdot  \circ a_r^{(N)}$, which are {\em rank} {\em one} tensors, and can be denoted as
\begin{equation}\label{}
\begin{array}{l}
{\cal X} \cong \sum\limits_{r = 1}^R {{\lambda _r}a_r^{(1)} \circ a_r^{(2)} \circ  \cdot  \cdot  \cdot  \circ a_r^{(N)}} \\
\;\;\;\; = \Lambda  \times {}_1{A^{(1)}} \times {}_2{A^{(2)}} \cdot  \cdot  \cdot  \times {}_N{A^{(N)}}\\
\end{array}
\end{equation}

Given $m$ views $\{X_p\}_{p=1}^m$ of samples, in which $X_p$ = \{$ \rm x_{p1}$, $ \rm x_{p2}$, ..., $\rm x_{pN}$ \}$\in$ $R^{d_p \times N}$, the variance matrices are ${C_{pp}} = \frac{1}{N}\sum\nolimits_{n = 1}^N {{{\mathop{\rm x}\nolimits} _{pn}}{\mathop{\rm x}\nolimits} _{pn}^T}$, $p$ = 1, 2, ..., $m$. Then, the covariance tensor among all of views is calculated as
\begin{equation}\label{}
{\cal C}_{1, 2, ..., m} = \frac{1}{N}\sum\nolimits_{n = 1}^N {{{\mathop{\rm x}\nolimits} _{1n}}
\circ {{\mathop{\rm x}\nolimits} _{2n}} \circ ... \circ {{\mathop{\rm x}\nolimits} _{mn}}}
\end{equation}
where $\cal C$ is a tensor with ${d_1} \times {d_2} \times ... \times {d_m}$. According to the traditional two-view CCA \cite{Hardoon2004Canonical}, exploration is performed to maximize the correlation among the canonical variables ${{\mathop{\rm z}\nolimits} _p} = X_p^T{{\mathop{\rm h}\nolimits} _p}$, $p$ = 1, 2, ..., $m$, in which $\{{{\mathop{\rm h}\nolimits} _p} \in {R^{{d_p} \times 1}}\}_{p=1}^m$ denotes the canonical vectors. Therefore, the optimization problem is
\begin{equation}\label{}
\begin{array}{l}
\mathop {\arg \max }\limits_{\{ {{\mathop{\rm h}\nolimits} _p}\} }  = {\mathop{\rm corr}\nolimits} ({{\mathop{\rm z}\nolimits} _1},{{\mathop{\rm z}\nolimits} _2},\;...,\;{{\mathop{\rm z}\nolimits} _m})\\
\;\;\;\;\;\;\;\;\;\;\;\;s.t.\;z_p^T{z_p} = 1,\;p = 1,\;...,\;m
\end{array},
\label{123}
\end{equation}

Here, ${\mathop{\rm corr}\nolimits} ({{\mathop{\rm z}\nolimits} _1},{{\mathop{\rm z}\nolimits} _2},\;...,\;{{\mathop{\rm z}\nolimits} _m}) = {({{\mathop{\rm z}\nolimits} _1} \odot {{\mathop{\rm z}\nolimits} _2} \odot ..., \odot {{\mathop{\rm z}\nolimits} _m})^T}{\rm e}$ expresses the canonical correlation, where $\bigodot$ denotes the element-wise product, and ${\mathop{\rm e}\nolimits}  \in {R^N}$. According to TCCA, the optimization problem (\ref{123}) is equivalent to
\begin{equation}\label{}
\begin{array}{l}
\mathop {\arg \max \;\rho }\limits_{\{ {{\mathop{\rm h}\nolimits} _p}\} }  = {{\cal C}_{1,2,\;...,\;m}}{\mathop  \times \limits^ -}_1{\mathop{\rm h}\nolimits} _1^T{\mathop  \times \limits^ -}_2{\mathop{\rm h}\nolimits} _2^T...{\mathop  \times \limits^-}_m{\mathop{\rm h}\nolimits} _m^T\\
\;\;\;\;\;\;\;\;\;\;\;s.t.\;\;{\mathop{\rm h}\nolimits} _p^T{C_{pp}}{{\mathop{\rm h}\nolimits} _p} = 1,\;p = 1,2,\;...,\;m
\end{array},
\label{124}
\end{equation}
where ${{\mathop\times \limits^ -}_p}$ denotes the $p$-mode contracted tensor-vector product. Let  ${{\mathop{\rm u}\nolimits} _p}$ = $\widetilde{C}_{pp}^{1/2}{\rm h}$ and ${\cal M} = {{\cal C}_{1,2,\;...,\;m}}$${\mathop  \times \limits^-}_1$${\widetilde{C}_{11}^{1/2}{\rm h}} {\mathop  \times \limits^ -}_2 \widetilde{C}_{22}^{1/2}{\rm h}\\...{\mathop  \times \limits^ -}_m \widetilde{C}_{mm}^{1/2}{\rm h}$. Then, the optimization problem in (\ref{124}) is described as
\begin{equation}\label{}
\begin{array}{l}
\mathop {\arg \max \;\rho }\limits_{\{ {{\mathop{\rm h}\nolimits} _p}\} }  = {\cal M}{\mathop  \times \limits^ -}_1{\mathop{\rm u}\nolimits} _1^T{\mathop  \times \limits^ -}_2{\mathop{\rm u}\nolimits} _2^T...{\mathop  \times \limits^ -}_m{\mathop{\rm u}\nolimits} _m^T\\
\;\;\;\;\;\;\;\;\;\;\;s.t.\;\;{\mathop{\rm u}\nolimits} _p^T{{\mathop{\rm u}\nolimits} _p} = 1,\;p = 1,2,\;...,\;m
\end{array},
\label{125}
 \end{equation}
where $\widetilde{C}_{pp}$ = $C_{pp}$ + $\epsilon I$, $\epsilon$ expresses a nonnegative trade-off parameter and $I$ denotes the identity matrix.

According to \cite{Lathauwer2006On}, Equation (\ref{125}) is equivalent to seeking the best rank-1 approximation of the tensor $\cal M$, {\em i.e.}, the best {\em rank} {\em one} CP decomposition of the tensor $\cal M$. This construct denoted as
\begin{equation}\label{}
{\cal M} \approx \sum\limits_{k = 1}^r {{\rho _k}} {\mathop{\rm u}\nolimits} _1^{(k)} \circ {\mathop{\rm u}\nolimits} _2^{(k)} \circ ... \circ {\mathop{\rm u}\nolimits} _p^{(k)},
\label{126}
\end{equation}

The alternating least squares (ALS) algorithm is used to approximately seek the solutions. Letting $U_p$ = $[{\rm u}_p^{(1)}, ..., {\rm u}_p^{(r)}]$, the projected data for the $p$'th view can be calculated as
\begin{equation}\label{}
{Z_p} = X_p^T\widetilde{C}_{pp}^{ - 1/2}{U_p}.
\end{equation}

The different ${Z_p}_{p=1}^m$ are concatenated as the final representation $Z \in {R^{(mr) \times N}}$ for the subsequent learning.
\section{Proposed Method}

%

\subsection{Low-level Feature Sharing for Face Attr-ibutes}

\begin{figure*}[t]
\setlength{\fboxsep}{0pt}
\setlength{\fboxrule}{0pt}
\begin{center}
\includegraphics[height =2.0in,width=5.5in]{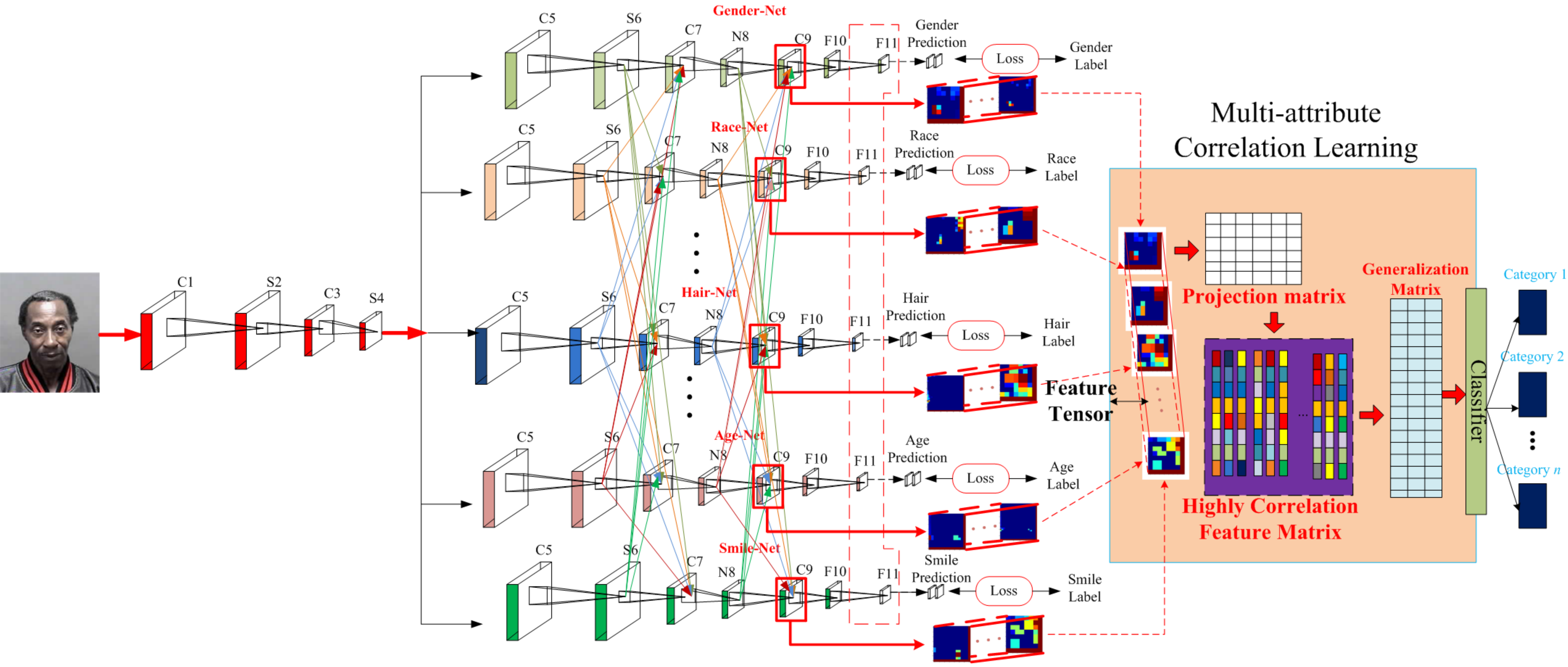} \\
\DeclareGraphicsExtensions.
\end{center}
\caption{Full schematic diagram of our network architecture. (C1, C3, ..., C9) denote the corresponding convolutional layers, (S2, S4, S6) represent pooling and normalization operations, N8 signifies only the normalization operation, and (F10 and F11) express the fully connected layers. The structure shares the information from C1 to S4 but splits that in high-level features (e.g., from the third convolutional layer to the fully connected layer). The feature maps of the C9 layers of the fine-tuned subnetworks are built into a feature tensor, and the tensor is projected into a highly correlated space via NTCCA, based on which the finial predictions are made.}
\label{fig:2}
\end{figure*}

The convolutional layers of a typical CNN model provide multiple levels of abstraction in the feature hierarchies \cite{ma2015hierarchical}. The features in the earlier layers retain higher spatial resolution for precise localization with low-level visual information. Because max pooling is used in the CNNs, the spatial resolution is gradually reduced with an increase in network depth. Therefore, the features in high-level layers capture more semantic information and fewer fine-grained spatial details. The face attributes ({\em e.g.,} lips, nose, hair) keep more semantic information than spatial resolution; in other words, the high-level features extracted from a face image are beneficial for face attribute prediction. Hence, for face multi-attribute prediction, the low-level features can be shared. According to \cite{ma2015hierarchical} and \cite{hand2017attributes}, because the first and second convolutional layers retain higher spatial resolution with low-level visual information, our MTCN shares low-level features from the input to the second convolutional layer. Fig. \ref{fig:2} shows a full schematic diagram of our network architecture.

\subsection{Differentiation and Correlation in High-level Layers }
From the third convolutional layer, we split the network into multi-subnetworks. This arrangement is chosen because different CNNs trained by different targets can be considered different feature descriptors, and the features learned from them can be seen as different views/representations of the data. These subnetworks have the same network structure and aim to predict different face attributes.

At the same time, based on \cite{7755833}, \cite{7060650}, \cite{5543609}, and \cite{8082561}, each of the face attribute estimations can be enhanced based on the other attributes, and each of our subnetworks seeks useful information from the other networks in the same layers to enhance itself. This operation appears twice in the C7 and C9 layers because these layers have more semantic information.

\begin{figure}[!hbpt]
\centering
\setlength{\fboxsep}{0pt}
\setlength{\fboxrule}{0pt}
\includegraphics[height =1.4in,width=2.8in]{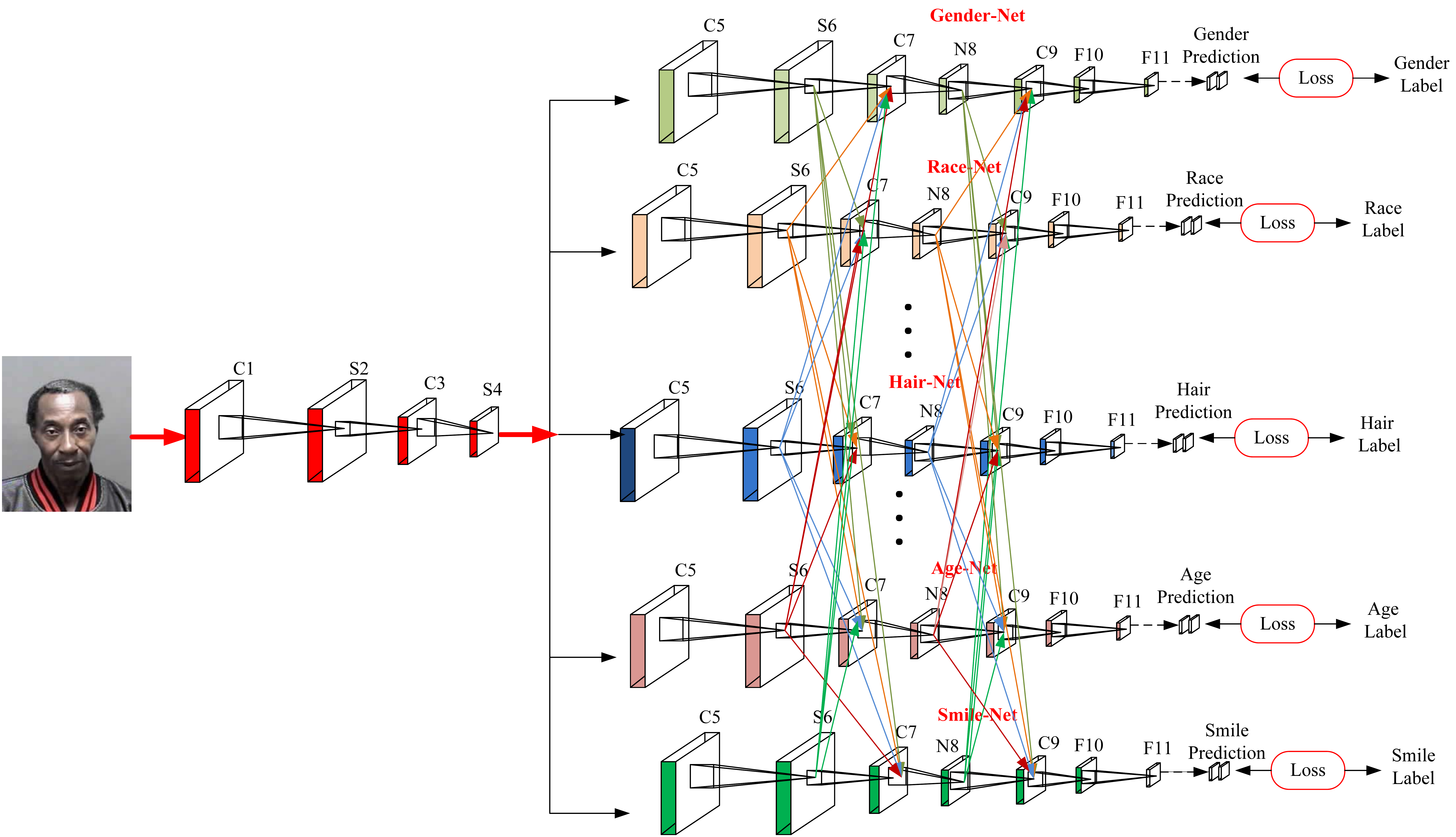} \\
\DeclareGraphicsExtensions.
\caption{The learning process of the subnetworks. }
\label{fig:3}
\end{figure}

In the first stage, as shown in Fig. \ref{fig:3}, the convolutional neural network is trained on datasets. In this situation, the whole structure is an end-to-end learning network. During the process of feed-forward processing, the low-level features are shared until the third convolutional layer and split at the high-level layers for task-specific losses.

Due to the specificity of the MTCN, backpropagation is a crucial step, and the gradients transferred from the output to C9, C9 to C7, and C5 to C3 are difficult to compute. We present the detailed derivations and the implementation in the following subsections. First, we use the cross-entropy loss function for the subnetworks, and the loss is
\begin{equation}\label{}
C =  - \frac{1}{N}\sum\limits_{i = 1}^N {\left( {{y_i}\ln {p_i} + (1 - {y_i})(1 - \ln {p_i})} \right)}.
\end{equation}
where $p_i$ denotes the probability of an attribute produced by our proposed network. We use $y_i$ to denote the ground-truth of the attribute and $N$ to denote the number of training examples.
\subsubsection{Gradients Transferred from the C9 layer to the N8 layer}
Our MTCN has two specific feature extraction stages, in which the convolutional layer extracts features from both its own network and from the same level layer of other subnetworks. For this reason, the operations in the whole subnetworks are the same in this stage, and we present only the detailed gradient transferred in Gender-Net. $K$ is the number of subnetworks. We assume that the weights and biases between C9 and the fully connected layer are $\textbf{w}_{cf}$ and $\textbf{b}_{cf}$ and that those between the N8 layer and C9 layer are $\textbf{w}_{nc}$ and $\textbf{b}_{nc}$. Here, $X_{c}^{i}$ and $X_{n}^{i}$ express the output of the convolutional and normalization layers of the $i$th sample. Although the C9 layer of Gender-Net extracts features from multiple subnetworks, we do not design other convolutional kernels for these feature maps. For example, ($X_1^i$, $X_2^i$, ..., $X_K^i$) denotes the corresponding feature maps of the $K$ subnetworks, and the outputs of the C9 layer of Gender-Net are
\begin{equation}\label{7}
X_{c}^i = f(X_1^i\textbf{w}_{nc} + X_2^i\textbf{w}_{nc}, ..., + X_K^i\textbf{w}_{nc} + \textbf{b}_{nc} ).
\end{equation}


Let us calculate the partial derivative of the cross-entropy cost with respect to the weights and biases. By applying the chain rule, we obtain
\begin{equation}\label{}
\frac{{\partial C}}{{\partial {w_{nc}}}} = \frac{{\partial C}}{{\partial {X_c^i}}}\frac{{\partial {X_c^i}}}{{\partial {w_{nc}}}},
\end{equation}
\begin{equation}\label{}
\frac{{\partial C}}{{\partial {b_{nc}}}} = \frac{{\partial C}}{{\partial X_c^i}}\frac{{\partial X_c^i}}{{\partial {b_{nc}}}},
\end{equation}
while
\begin{equation}\label{}
\frac{{\partial C}}{{\partial {X_c^i}}} =  - \frac{1}{N}\sum\limits_{i = 1}^N {\frac{{(y - f({w_{cf}},{X_c^i},{b_{cf}})){f^{'}}({w_{cf}},{X_c^i},{b_{cf}}){w_{cf}}}}{{f({w_{cf}},{X_c}^i,{b_{cf}})(1 - f({w_{cf}},{X_c}^i,{b_{cf}}))}}},
\end{equation}

We use the definition of the ReLU function, $max(0, x)$, and then $f'(x) = \left\{ \begin{array}{l}
1\;\;\;x > 0\\
0\;\;\;x \le 0
\end{array} \right.$, where $x = \sum\limits_i {{w_{cf}}X_c^i}  + {b_{cf}}$. Thus,
\begin{equation}\label{}
\frac{{\partial C}}{{\partial X_c^i}} = \left\{ \begin{array}{l}
\frac{1}{N}\sum\limits_{i = 1}^N {{w_{cf}}} \frac{{\left( {y - f({w_{cf}},X_c^i,{b_{cf}})} \right)}}{{f({w_{cf}},X_c^i,{b_{cf}})\left( {f({w_{cf}},X_c^i,{b_{cf}}) - 1} \right)}}\;\;x > 0\\
0\;\;\;\;\;\;\;\;\;\;\;\;\;\;\;\;\;\;\;\;\;\;\;\;\;\;\;\;\;\;\;\;\;\;\;\;\;\;\;\;\;\;\;\;\;\;\;\;\;\;\;\;\;\;\;\;\;\;\;\;x \le 0
\end{array} \right.\;\;\;,
\end{equation}
and according to equation (\ref{7}),
\begin{equation}\label{}
\frac{{\partial X_c^i}}{{\partial {w_{nc}}}} = f'({w_{nc}},X_n^i,{b_{nc}})\sum\limits_{j = 1}^K {X_j^i}  = \left\{ \begin{array}{l}
\sum\limits_{j = 1}^K {X_j^i} \;\;x > 0\\
0\;\;\;\;\;\;\;\;\;\;x \le 0
\end{array} \right.,
\end{equation}
\begin{equation}\label{}
\frac{{\partial X_c^i}}{{\partial {b_{nc}}}} = f'({w_{nc}},X_n^i,{b_{nc}}) = \left\{ \begin{array}{l}
1\;\;\;x > 0\\
0\;\;\;x \le 0
\end{array} \right.,
\end{equation}

Therefore,
\begin{equation}\label{}
\frac{{\partial C}}{{\partial {w_{nc}}}} = \left\{ \begin{array}{l}
\frac{1}{N}\sum\limits_{i = 1}^N {\sum\limits_{j = 1}^K {\frac{{\left( {y - f({w_{cf}},X_c^i,{b_c})} \right){w_{cf}}X_j^i\;}}{{f({w_{nc}},X_n^i,{b_{nc}})\left( {{f}({w_{nc}},X_n^i,{b_{nc}}) - 1} \right)}}} } \;x > 0\;\\
0\;\;\;\;\;\;\;\;\;\;\;\;\;\;\;\;\;\;\;\;\;\;\;\;\;\;\;\;\;\;\;\;\;\;\;\;\;\;\;\;\;\;\;\;\;\;\;\;\;\;\;\;\;\;\;\;\;\;\;\;\;x \le 0
\end{array} \right.,
\end{equation}
\begin{equation}\label{}
\frac{{\partial C}}{{\partial {b_{nc}}}} = \left\{ \begin{array}{l}
\frac{1}{N}\sum\limits_{i = 1}^N {\frac{{\left( {y - f({w_{cf}},X_c^i,{b_c})} \right){w_{cf}}\;}}{{f({w_{nc}},X_n^i,{b_{nc}})\left( {f({w_{nc}},X_n^i,{b_{nc}}) - 1} \right)}}} \;\;x > 0\;\\
0\;\;\;\;\;\;\;\;\;\;\;\;\;\;\;\;\;\;\;\;\;\;\;\;\;\;\;\;\;\;\;\;\;\;\;\;\;\;\;\;\;\;\;\;\;\;\;\;\;\;\;\;\;\;\;\;x \le 0
\end{array} \right.,
\end{equation}
and we can update the weights and biases in this layer as follows:
\begin{equation}\label{}
{w_{nc}} = {w_{nc}} + \eta \frac{{\partial C}}{{\partial {w_{nc}}}},
\end{equation}
\begin{equation}\label{}
{b_{nc}} = {b_{nc}} + \eta \frac{{\partial C}}{{\partial {b_{nc}}}}.
\end{equation}
where $\eta$ is the learning rate.
\subsubsection{Gradients Transferred from the N8 layer to the S6 layer}
The partial derivative of the cross-entropy cost with respect to the weights and biases from the C7 layer to the S6 layer is the same as that from the C9 layer to the N8 layer. It is important to consider how to transfer the gradients from the N8 layer to the C7 layer in Gender-Net because the C9 layer extracts features from multiple subnetworks; these features affect the gradients simultaneously because Gender-Net is a full subnetwork. In this time, $\textbf{w}_{sc}^{'}$ and $\textbf{b}_{sc}^{'}$ signify the weights and biases between the S6 layer and the C7 layer, respectively, and $X_c^{i'}$ denotes the output of the C7 layer. We apply the chain rule twice to compute the partial derivative as follows:
\begin{equation}\label{}
\frac{{\partial C}}{{\partial w_{sc}^{'}}} = \frac{{\partial C}}{{\partial X_c^i}}\frac{{\partial X_c^i}}{{\partial X_c^{i'}}}\frac{{\partial X_c^{i'}}}{{\partial w_{sc}^{'}}},
\end{equation}
\begin{equation}\label{}
\frac{{\partial C}}{{\partial b_{sc}^{'}}} = \frac{{\partial C}}{{\partial X_c^i}}\frac{{\partial X_c^i}}{{\partial X_c^{i'}}}\frac{{\partial X_c^{i'}}}{{\partial b_{sc}^{'}}},
\end{equation}
due to $X_c^i = f(X_c^{i'}{w_{sc}} + \;...\; + {b_{sc}})$ and $f'(x) = \left\{ \begin{array}{l}
1\;\;\;x > 0\\
0\;\;\;x \le 0
\end{array} \right.$, where $x = \sum\limits_i {{w_{sc}}X_s^i}  + {b_{sc}}$. We can find
\begin{equation}\label{}
\frac{{\partial X_c^i}}{{\partial X_c^{i'}}} = f'\left( {{w_{sc}},X_s^i,{b_{sc}}} \right){w_{sc}} = \left\{ \begin{array}{l}
{w_{sc}}\;\;x > 0\\
0\;\;\;\;\;\;x \le 0
\end{array} \right.,
\end{equation}
and
\begin{equation}\label{}
\frac{{\partial X_c^{i'}}}{{\partial w_{sc}^{'}}} = f'(w_{sc}^{'},X_s^{i'},b_{sc}^{'})\sum\limits_{j = 1}^{K'} {X_j^i} ' = \left\{ \begin{array}{l}
\sum\limits_{j = 1}^{K'} {X_j^i} '\;\;x > 0\\
0\;\;\;\;\;\;\;\;\;\;\;\;x \le 0
\end{array} \right.,
\end{equation}
\begin{equation}\label{}
\frac{{\partial X_c^{i'}}}{{\partial b_{sc}^{'}}} = f'(w_{sc}^{'},X_s^{i'},b_{sc}^{'}) = \left\{ \begin{array}{l}
1\;\;\;\;x > 0\\
0\;\;\;\;x \le 0
\end{array} \right.,
\end{equation}

Therefore,
\begin{equation}\label{}
\frac{{\partial C}}{{\partial {w_{sc}}}} = \left\{ \begin{array}{l}
\frac{1}{N}\sum\limits_{i = 1}^N {\sum\limits_{j = 1}^{K'} {\frac{{\left( {y - f({w_{cf}},X_c^i,{b_c})} \right){w_{cf}}{w_{sc}}X_j^{i'}\;}}{{f({w_{sc}},X_s^i,{b_{sc}})\left( {f({w_{sc}},X_s^i,{b_{sc}}) - 1} \right)}}} } \;x > 0\;\\
0\;\;\;\;\;\;\;\;\;\;\;\;\;\;\;\;\;\;\;\;\;\;\;\;\;\;\;\;\;\;\;\;\;\;\;\;\;\;\;\;\;\;\;\;\;\;\;\;\;\;\;\;\;\;\;\;\;\;\;x \le 0
\end{array} \right.,
\end{equation}
\begin{equation}\label{}
\frac{{\partial C}}{{\partial b_{sc}^{'}}} = \left\{ \begin{array}{l}
\frac{1}{N}\sum\limits_{i = 1}^N {\frac{{\left( {y - f({w_{cf}},X_c^i,{b_c})} \right){w_{cf}}{w_{sc}}\;}}{{f({w_{sc}},X_s^i,{b_{sc}})\left( {f({w_{sc}},X_s^i,{b_{sc}}) - 1} \right)}}} \;\;x > 0\;\\
0\;\;\;\;\;\;\;\;\;\;\;\;\;\;\;\;\;\;\;\;\;\;\;\;\;\;\;\;\;\;\;\;\;\;\;\;\;\;\;\;\;\;\;\;\;\;\;\;\;\;\;\;\;\;x \le 0
\end{array} \right.,
\end{equation}

Then, the weights and biases between the S6 layer and the C7 layer can be updated as
\begin{equation}\label{}
{w_{sc}^{'}} = {w_{sc}^{'}} + \eta \frac{{\partial C}}{{\partial {w_{sc}^{'}}}},
\end{equation}
\begin{equation}\label{}
{b_{sc}^{'}} = {b_{sc}^{'}} + \eta \frac{{\partial C}}{{\partial {b_{sc}^{'}}}}.
\end{equation}
\subsubsection{Gradients Transferred from Subnetworks to a Shared Single Network}
The weights and biases between the S4 and C5 layers can be updated based on the corresponding subnetworks. How to transfer the gradients from the subnetworks to a single network is another crucial problem. Due to the distinctiveness of our MTCN, we adopt a joint gradient transfer strategy to compute the gradients. {$C_1$, $C_2$, ..., $C_K$} denote the cross-entropy losses of the whole subnetworks. Additionally, $\textbf{w}_{sc}^{''}$ and $\textbf{b}_{sc}^{''}$ express the weights and biases between the S4 layer and the C5 layer, and $\textbf{w}_{sc}^{'''}$ and $\textbf{b}_{sc}^{'''}$ signify the weights and biases between the S2 layer and the C3 layer. The joint gradient transferred strategy is
\begin{equation}\label{}
\Delta w = \frac{{\partial {C_1}}}{{\partial w_{sc}^{'''}}} + \frac{{\partial {C_2}}}{{\partial w_{sc}^{'''}}} + \;...\; + \;\frac{{\partial {C_K}}}{{\partial w_{sc}^{'''}}},
\end{equation}
\begin{equation}\label{}
\Delta b = \frac{{\partial {C_1}}}{{\partial b_{sc}^{'''}}} + \frac{{\partial {C_2}}}{{\partial b_{sc}^{'''}}} + \;...\; + \;\frac{{\partial {C_K}}}{{\partial b_{sc}^{'''}}},
\end{equation}
where $\frac{{\partial {C_1}}}{{\partial w_{sc}^{'''}}}$, ..., $\frac{{\partial {C_K}}}{{\partial w_{sc}^{'''}}}$ and $\frac{{\partial {C_1}}}{{\partial b_{sc}^{'''}}}$, ..., $\frac{{\partial {C_K}}}{{\partial b_{sc}^{'''}}}$ can be calculated by the chain rule based on the calculation of the partial derivatives above.

Therefore, we can update the weights and biases between the S2 layer and the C3 layer as
\begin{equation}\label{}
w_{sc}^{'''} = w_{sc}^{'''} + \eta \Delta w,
\end{equation}
\begin{equation}\label{}
b_{sc}^{'''} = b_{sc}^{'''} + \eta \Delta b.
\end{equation}

\subsection{Multi-attribute Tensor Correlation Learning Framework}
In the first stage, the subnetworks not only consider the differences among them but also extract the related information to enhance themselves. Although this novel design can achieve better performance than can most of the compared methods, we do not fully consider the specific degrees of correlation among the face attributes. Hence, based on the fine-tuned network, we want to further excavate the detailed correlation information, so a novel TCCA approach (called NTCCA) is proposed to explore the detailed correlations among the high-level features of these subnetworks. Unlike TCCA, which aims to directly maximize the correlation between the canonical variables of all views \cite {Luo2015Tensor}, our proposed NTCCA maximizes the correlation of all of the feature maps in C9 for an image.

To explore the correlation among the different types of features in C9 for an image, we consider ${X_l^i}$ = $\{\{X_1^1, X_2^1,$ ..., $X_L^1\}$, $\{X_1^2, X_2^2, ..., X_L^2\}$, ..., $\{X_1^K, X_2^K, ..., X_L^K\}\}$, $l$ = 1, 2, ..., $L$ and $i$ = 1, 2, ..., $K$. The size of the feature map in C9 is $\kappa$$\times$$\kappa$, and ${X_l^i}$ composes a 3-D tensor, $\cal{X}$ $\in$ $R^{\kappa\times\kappa\times KL}$ where $KL$ denotes the whole feature maps. Based on $\cal{X}$, we redefine the feature map as $\{X_p\}_{p=1}^{KL}$ and ${X_p}$ =$\{x_{p1}, x_{p2}, ..., x_{p\kappa}\} \in R^{\kappa\times\kappa}$. The variance matrices can be denoted as ${ C_{dd}} = \frac{1}{{{\kappa}}}\sum\nolimits_{j = 1}^{{k}} {{x_{pj}}x_{pj}^T}$, and the covariance tensor among $X_{1}$, $X_{2}$, ..., $X_{KL}$ is calculated as

\begin{equation}\label{414}
\begin{array}{l}
{ {\cal C}_{1,2, ..., (KL)}}\; = \;\frac{1}{\kappa}\sum\nolimits_{j = 1}^{\kappa} {{x_{1j}}}  \circ {x_{2j}} \circ \cdot \cdot \cdot \circ {x_{(KL)j}},
\end{array}
\end{equation}
where ${\cal C}$ is a tensor of dimension ${\kappa} \times {\kappa} \times \cdot \cdot \cdot \times {\kappa}$ and $\circ$ expresses the outer product.

Without loss of generality, we first obtain the canonical correlation as Equation (\ref{4}), where the canonical variables ${z_p} = X_p^T{h_p}$.
\begin{equation}\label{4}
\begin{array}{l}
\arg \max \rho  = corr({z_1},{z_2}, \; \cdot  \cdot  \cdot, \;\mathop  {z_{KL}})\\
\\
\;\;\;\;\;\;\;\;\;\;\;\;\;\;s.t.\;z_p^T{z_p} = 1,\;p = 1,\;2,\; \cdot  \cdot  \cdot, \;(KL),
\end{array}
\end{equation}

According to TCCA, Equation (\ref{4}) is equivalent to ${C_{1,2,\;...,\;(KL)}}\\ \mathop  \times \limits^ -  {}_1h_1^T\mathop  \times \limits^ -  {}_2h_2^T\mathop  \times \limits^ -  \; \cdot  \cdot  \cdot \;\mathop  \times \limits^ -  {}_{(KL)}h_{(KL)}^T$, and the correlation can be further calculated as
\begin{equation}\label{5}
\begin{array}{l}
\arg \max \rho  = {C_{1,2,\;...,\;(KL)}}\mathop  \times \limits^ -  {}_1h_1^T\mathop  \times \limits^ -  \; \cdot  \cdot  \cdot \;\mathop  \times \limits^ -  {}_{(KL)}h_{(KL)}^T \\
\\
\;\;\;\;\;\;\;\;\;\;\;\;\;\;\;\;\;\;\;\;\;\;\;s.t.\;h_p^T{C_{pp}}{h_p} = 1,
\end{array}
\end{equation}
where ${X_p}X_p^T = {C_{pp}}$ and $\mathop  \times \limits^ -  {}_p$ denote the $p$-mode contracted tensor-vector product.

According to Equations (\ref{125}) and (\ref{126}), the alternating least squares (ALS) algorithm is used to seek approximate solutions. Letting $U_p$ = $[{\rm u}_p^{(1)}, ..., {\rm u}_p^{(r)}]$, the projected data for the $p$'th view can be calculated as
\begin{equation}\label{115}
{Z_p} = X_p^T\widetilde{C}_{pp}^{ - 1/2}{U_p}.
\end{equation}

Then, we concatenate the different $\{Z_p\}_{p=1}^{(KL)}$ as the final representation $Z \in R^{(KLr)\times \kappa}$. Because the method presented above is only used to calculate the correlation of multiple attributes of an image, a  generalization matrix is utilized to ensure that the projected results exhibit more stabilization and higher correlation. Parts of the training dataset are used to train the matrix through algorithm \ref{alg:1111}. Our goal is to estimate multiple attributes for an image; thus, a joint attribute estimation model is utilized to calculate the loss of the whole system. For a face image with $M$ attributes, a joint attribute estimation model can be formulated as follows:
\begin{equation}\label{358}
\epsilon = \arg \;min\sum\limits_{i = 1}^M {{C_i}}  + \gamma \Phi (W_j).
\end{equation}
where $C_i$ expresses the cross-entropy loss of the $i$th attribute, $\Phi(\cdot)$ denotes a regularization term to penalize the complexity of the weights, and $\gamma > 0$ is a regularization parameter.

During this process, the neural network is not updated, and we only update $W$ and $b$. Algorithm \ref{alg:1111} is as follows:

\begin{algorithm}[t]
\caption{Novel Tensor Canonical Correlation Analysis}
\begin{algorithmic}[1]
\REQUIRE ~~\\
The training set: $N$ face images;\\
Iterations $t$ and max iterations $t_{max}$;\\
Error $\epsilon$ and minimum error $e_{min}$; Learn rate $\eta$.\\
\ENSURE ~~\\
Output layer weight and bias: $W$ and $b$.
\STATE Initialize the output layer weight and bias: $W$ and $b$;
\STATE \textbf{for} $i$ in range ($N$)
\STATE \quad\ Map $KL$ attribute feature space into another space according to Equation (\ref{115}); \\
\quad\ \{$X_1, X_2, ..., X_{(KL)}$\} $\rightarrow$ \{$Z_1$, $Z_2$, ..., $Z_{(KL)}$ \};
\STATE \quad\ Calculate muli-attribute output: $y$ = $Z$ $\cdot$ $W$ + $b$;
\STATE \quad\ Calculate final total loss $\epsilon$ according to Equation (\ref{358});
\STATE \quad\   \textbf{if} ($\epsilon$ $\leq$ $e_{min}$) or ($t$ $\geq$ $t_{max}$)
\STATE \qquad\    Use the fine-tuned model to predict the multi-attribute tasks;
\STATE \quad\   \textbf{else} Compute the modified weight coefficient:
\STATE       \qquad\        $\Delta w = \eta \frac{{\partial \epsilon}}{{\partial w}}$;\\
\STATE \qquad\    Compute the modified biases coefficient: \\
       \qquad\    $\Delta b = \eta \frac{{\partial \kappa }}{{\partial b}}$;\\
\STATE \qquad\ Update output weights $W = W + \Delta w$;
\STATE \qquad\ Update output biases $b = b + \Delta b$;
\STATE \quad\ \textbf{end if}
\STATE  \textbf{end for}
\STATE Return updated output layer weight $W$ and bias $b$.
\end{algorithmic}
\label{alg:1111}
\end{algorithm}

CelebA and LFWA datasets are used in our experiments \cite{7410782} and they are divided into training dataset, validation dataset, and testing dataset. Till now, our MTCN has been fine-tuned and the training process are roughly as follows:\\
Step 1: Train MTCN without NTCCA on the training datasets with the corresponding attributes and a fine-tuned MTCN can be used to make predictions;\\
Step 2: Train the generalization matrix with NTCCA on one third of the training datasets;\\
Step 3: Verify the performance of the fine-tuned MTCN on the testing datasets.\\

\section{Experiments}
\subsection{Datasets}
\subsubsection{CelebA}
CelebA is a large-scale face attribute database \cite{7410782}; it contains 10K identities, and each identity has 20 images. Each image has 40 attributes (see Table \ref{tab:1}), such as gender, race, and smiling, which makes it challenging for face attribute prediction. The dataset contains 200,000 images: 160,000 are used for training, 20,000 for validation, and 20,000 for testing. Because the CelebA dataset is so large, we do not need to augment it in any way.

\begin{table}[h]
\centering
\caption{Summary of the 40 face attributes provided in the CelebA dataset.}
\scalebox{0.65}[0.6]{
\begin{tabular}{c|c|c|c}
\toprule[2pt]
 Attr. Idx.    & Attr. Def     &Attr. Idx  &Attr. Def \\ \midrule[1pt]
 1 & 5 O'ClockShadow & 21 &Male\\
 2& ArchedEyebrows & 22&MouthSlighlyOpen\\
 3&BushyEyebrows&23&Mustache\\
 4&Attractive&24&NarrowEyes\\
 5&BagsUnderEyes&25&NoBeard\\
 6&Bald&26&OvalFace\\
 7&Bangs&27&PaleSkin\\
 8&BlackHair&28&PointyNose\\
 9&BlondHair&29&RecedingHairline\\
 10&BrownHair&30&RosyCheeks\\
 11&GrayHair&31&SideBurns\\
 12&BigLips&32&Smiling\\
 13&BigNose&33&StrightHair\\
 14&Blurry&34&WavyHair\\
 15&Chubby&35&WearEarrings\\
 16&DoubleChin&36&WearHat\\
 17&Eyeglasses&37&WearLipstick\\
 18&Goatee&38&WearNecklace\\
 19&HeavyMakeup&39&WearNecktie\\
 20&HighCheekbones&40&Young \\
\bottomrule[2pt]
\end{tabular}}
\label{tab:1}
\end{table}
\subsubsection{LFWA}
LFWA is another unconstrained face attribute database \cite{7410782}, and its face images are from the LFW database \cite{Huang2007Labeled}. It has 40 attributes, which have the same annotations as in the CelebA database. The LFWA dataset consists of 13,143 images, of which, 6,263 were used for training, 2,800 for validation, and 4,080 for testing. If we did not augment the training dataset, then the network would have severely overfit the dataset because of the large number of parameters. We follow the data augmentation scheme presented in \cite{hand2017attributes} and we have over 75,000 images for training.
\subsection{Implementation Details}
 Our proposed structure is implemented using the publicly available Tensorflow \cite{Abadi2016TensorFlow} code. The entire network in this paper is trained using an NVIDIA Tesla P100. First, we resize the input image to 256 $\times$ 256 pixels, and then, a 224 $\times$ 224 crop is selected from the center of the image or the four corners from the entire processed image. We also adopt different dropout measures to limit the risk of overfitting. The network is initialized with random weights following a Gaussian distribution; the mean is 0, and the standard deviation is 0.01. A base learning rate of $10^{-4}$ is used, and it is reduced by 10\% every 100,000 iterations. To train the MTCN, we use batches of size 100, and we train both datasets for 30 epochs. Overall, training with NTCCA takes approximately 10 hours for the CelebA dataset and approximately 4 hours for the LFWA dataset, and nearly 1.5 hours is required to calculate the generalization matrix $W$. Each experiment is conducted four times, and we obtain the average of the relevant results. Because codes of the baseline methods used in subsequent sections are not available in the public domain, we directly report the results in the corresponding publications.
\begin{table}[!hbpt]
\centering
\caption{Subnetwork Parameters.}
\scalebox{0.65}[0.65]{%
\begin{tabular}{c|c|c|c|c|c}
\toprule[2pt]
\textbf{Layers} & \textbf{Parameters}&\textbf{Layers}&\textbf{Parameters}&\textbf{Layers}&\textbf{Parameters} \\\midrule[1pt]
\multirow{3}*{Conv1} & Num\_output: 96 &            & Num\_output: 96  &  & Local\_size: 5\\
     & Kernel\_size: 5 &    Pool1   & Kernel\_size: 3 &    Norm1    & alpha: 1e-1\\
      & Stride: 2 &                   & Stride: 2     & & beta: 0.75\\
\hline
 & Num\_output: 256 &            & Num\_output: 256  &  & Local\_size: 5\\
 Conv2      & Kernel\_size: 3 &    Pool2   & Kernel\_size: 3 &    Norm2    & alpha: 1e-1\\
      & Stride: 1&                   & Stride: 2     & & beta: 0.75\\
\hline
 & Num\_output: 384 &            & Num\_output: 384  &  & Local\_size: 5\\
Conv3      & Kernel\_size: 3 &    Pool3   & Kernel\_size: 3 &    Norm3    & alpha: 1e-1\\
      & pad: 1 &                   & Stride: 2     & & beta: 0.75\\
\hline
 & Num\_output: 384 &               & Local\_size: 5 & &Num\_output: 256\\
Conv4      & Kernel\_size: 3 &         Norm4    & alpha: 0.01&Conv5& Kernel\_size: 3\\
      & Stride: 1 &                     & beta: 0.75&&Stride: 1\\
\bottomrule[2pt]
\end{tabular}}
\label{tab:021}
\end{table}
\subsubsection{Network Structure}
 The neural network of the MTCN consists of two parts: the shared network and 40 subnetworks. The 40 subnetworks have the same network layers, such as convolutional layers, contrast normalization layer, pooling layer, ReLU nonlinear function, and identical network parameters. The detailed subnetwork configurations are shown in Table \ref{tab:021}. The convolutional layer is followed by ReLU, which is a max pooling and a local response normalization layer. Every F10 layer has 4098 units and is followed by a ReLU and 50\% dropout to avoid overfitting. Each F11 layer is fully connected to a corresponding F10 layer, which also has 4098 units, and it is also followed by ReLU and a 50\% dropout. The final fully connected layer connects F11 with 1000 units.

\begin{table*}[t]
\centering
\caption{Attribute estimation accuracies (in \%) for the 40 binary attributes (see Table 2) on the CelebA and LFWA
databases by the proposed approach and the state-of-the-art methods \cite{7410782}, \cite{rudd2016moon}, \cite{hand2017attributes}, \cite{Han2017Heterogeneous}, \cite{gunther2016affact}, and \cite{ding2017deep}. The average
accuracies of \cite{7410782}, \cite{rudd2016moon}, \cite{hand2017attributes}, \cite{Han2017Heterogeneous}, \cite{gunther2016affact}, \cite{gunther2016affact}(unaligned), \cite{ding2017deep}, and the proposed approach are 87.30\%, 90.94\%, 91.29\%, 92.60\%, 91.01\%, 90.32\%, 91.23\%, 91.95\%(Ours), and 92.97\%(Ours), respectively, on CelebA, and the average accuracies of \cite{7410782}, \cite{hand2017attributes}, \cite{Han2017Heterogeneous}, and the proposed approach are 83.85\%, 86.31\%, 86.15\%, 86.81\%(Ours), and 87.96\%(Ours), respectively, on LFWA. The highest accuracy for each attribute is in bold.}
\scalebox{0.6}[0.65]{
\begin{tabular}{cc|cccccccccccccccccccc}
\toprule[2pt]
\multicolumn{2}{c|}{\multirow{2}*{\textbf{Approach}}} & \multicolumn{20}{|c}{\textbf{Attribute index}} \\
               & &1&2&3&4&5&6&7&8&9&10&11&12&13&14&15&16&17&18&19&20\\\midrule[1pt]
\multirow{9}*{\rotatebox{90}{\textbf{CelebA}}}&LENet+ANet \cite{7410782}&84.00&82.00&83.00&83.00&88.00&88.00&75.00&81.00&90.00&\textbf{97.00}&74.00&77.00&82.00&73.00&78.00&95.00&78.00&84.00&\textbf{95.00}&88.00\\
&MOON \cite{rudd2016moon} &94.03&82.26&81.67&84.92&98.77&95.80&71.48&84.00&89.40&95.86&95.67&\textbf{89.38}&92.62&95.44&96.32&99.47&97.04&98.10&90.99&87.01   \\
&MCNN+AUX \cite{hand2017attributes} &94.51&83.42&83.06&84.92&98.90&96.05&71.47&84.53&89.78&96.01&96.17&89.15&\textbf{92.84}&95.67&96.32&99.63&97.24&98.20&91.55&87.58  \\
&DMTL \cite{Han2017Heterogeneous}&95.00&86.00&85.00&85.00&99.00&99.00&\textbf{96.00}&85.00&91.00&96.00&96.00&88.00&92.00&96.00&97.00&99.00&\textbf{99.00}&98.00&92.00&88.00  \\
&AFFACT \cite{gunther2016affact} &94.21&82.12&82.83&83.75&99.06&96.05&70.88&83.82&90.32&96.07&95.50&89.16&92.41&94.41&96.18&99.61&97.31&98.28&91.10&86.88  \\
&AFFACT Unaligned \cite{gunther2016affact} &94.09&81.27&80.36&84.89&97.82&95.49&71.42&81.83&85.88&95.17&94.52&87.72&90.59&95.10&95.94&99.38&97.21&97.89&90.82&86.11  \\
&PaW \cite{ding2017deep} &94.64&83.01&82.86&84.58&98.93&95.93&71.46&83.63&89.84&95.85&96.11&88.50&92.62&95.46&96.26&99.59&97.38&98.21&91.53&87.44  \\
&MTCN without NTCCA  &94.68&84.92&84.71&85.11&98.05&97.73&86.04&84.18&90.42&95.47&95.13&88.48&91.37&95.49&96.18&99.03&98.42&98.10&91.47&87.19  \\
&MTCN with NTCCA &\textbf{95.46}&\textbf{86.02}&\textbf{86.23}&\textbf{85.97}&\textbf{99.12}&\textbf{99.42}&95.44&\textbf{86.03}&\textbf{91.14}&96.82&\textbf{96.44}&89.28&92.00&\textbf{96.32}&\textbf{97.16}&\textbf{99.68}&98.73&\textbf{98.59}&92.34&\textbf{88.95}  \\ \midrule[1pt]
\multirow{6}*{\rotatebox{90}{\textbf{LFWA}}}&LENet+ANet \cite{7410782}&\textbf{84.00}&82.00&83.00&83.00&88.00&88.00&75.00&81.00&90.00&97.00&74.00&77.00&82.00&73.00&78.00&\textbf{95.00}&78.00&84.00&95.00&88.00\\
&MCNN+AUX \cite{hand2017attributes} &77.06&81.78&80.31&83.48&91.94&90.08&79.24&84.98&92.63&97.41&85.23&80.85&\textbf{84.97}&76.86&81.52&91.30&82.97&88.93&95.85&88.38  \\
&DMTL \cite{Han2017Heterogeneous}&80.00&86.00&82.00&84.00&92.00&93.00&77.00&83.00&92.00&97.00&89.00&81.00&80.00&75.00&78.00&92.00&86.00&88.00&95.00&89.00  \\
&MTCN without NTCCA  &80.59&85.14&82.35&83.78&92.01&92.78&80.64&84.51&92.17&97.28&87.97&80.91&83.00&79.01&80.24&91.67&85.58&88.74&95.72&88.63  \\
&MTCN with NTCCA &81.68&\textbf{86.23}&\textbf{83.01}&\textbf{84.33}&\textbf{92.16}&\textbf{93.44}&\textbf{84.51}&\textbf{85.17}&\textbf{93.20}&\textbf{98.09}&\textbf{89.47}&\textbf{81.83}&84.52&\textbf{83.27}&\textbf{82.00}&92.84&\textbf{87.12}&\textbf{89.81}&\textbf{96.41}&\textbf{89.75}  \\ \midrule[1pt]
\multicolumn{2}{c|}{\multirow{2}*{\textbf{Approach}}} & \multicolumn{20}{|c}{\textbf{Attribute index}} \\
               & &21&22&23&24&25&26&27&28&29&30&31&32&33&34&35&36&37&38&39&40\\\midrule[1pt]
\multirow{9}*{\rotatebox{90}{\textbf{CelebA}}}&LENet+ANet \cite{7410782}&94.00&82.00&92.00&81.00&79.00&74.00&84.00&\textbf{80.00}&85.00&78.00&77.00&91.00&76.00&76.00&\textbf{94.00}&88.00&\textbf{95.00}&88.00&79.00&86.00\\
&MOON \cite{rudd2016moon} &98.10&93.54&96.82&86.52&95.58&75.73&97.00&76.46&93.56&94.82&97.59&92.60&82.26&82.47&89.60&98.95&93.93&87.04&96.63&88.08  \\
&MCNN+AUX \cite{hand2017attributes} &98.17&93.74&96.88&87.23&96.05&75.84&97.05&77.47&93.81&95.16&97.85&92.73&83.58&83.91&90.43&99.05&94.11&86.63&96.51&88.48  \\
&DMTL \cite{Han2017Heterogeneous}&98.00&94.00&97.00&\textbf{90.00}&97.00&78.00&97.00&78.00&94.00&\textbf{96.00}&98.00&\textbf{94.00}&85.00&\textbf{87.00}&91.00&99.00&93.00&89.00&97.00&90.00  \\
&AFFACT \cite{gunther2016affact} &98.26&92.60&96.89&87.23&95.99&75.79&97.04&74.83&93.29&94.45&97.83&91.77&84.10&85.65&90.20&99.02&91.69&87.85&96.90&88.66  \\
&AFFACT Unaligned \cite{gunther2016affact} &97.29&92.82&96.89&87.15&95.33&74.87&96.97&76.24&91.74&94.54&97.46&90.45&82.17&83.37&90.33&98.66&92.99&87.55&96.43&86.21  \\
&PaW \cite{ding2017deep} &98.39&94.05&96.90&87.56&96.22&75.03&97.08&77.35&93.44&95.07&97.64&92.73&83.52&84.07&89.93&99.02&94.24&87.70&96.85&88.59  \\
&MTCN without NTCCA  &98.43&93.89&96.59&88.97&96.71&76.35&97.04&77.81&93.92&95.78&97.91&93.07&84.98&86.54&90.17&98.91&93.18&88.76&97.00&89.95  \\
&MTCN with NTCCA &\textbf{98.52}&\textbf{94.61}&\textbf{97.18}&89.42&\textbf{97.31}&\textbf{78.52}&\textbf{97.18}&78.47&\textbf{94.35}&\textbf{96.00}&\textbf{98.34}&93.91&\textbf{85.49}&\textbf{87.00}&91.04&\textbf{99.10}&94.00&\textbf{89.31}&\textbf{97.26}&\textbf{90.71}  \\ \midrule[1pt]
\multirow{6}*{\rotatebox{90}{\textbf{LFWA}}}&LENet+ANet \cite{7410782}&94.00&82.00&92.00&81.00&79.00&74.00&84.00&80.00&85.00&78.00&77.00&91.00&76.00&76.00&94.00&88.00&95.00&88.00&79.00&86.00\\
&MCNN+AUX \cite{hand2017attributes} &94.02&83.51&93.43&82.86&82.15&77.39&93.32&84.14&86.25&87.92&83.13&91.83&78.53&\textbf{81.61}&94.95&90.07&95.04&89.94&80.66&85.84  \\
&DMTL \cite{Han2017Heterogeneous}&93.00&86.00&95.00&82.00&81.00&75.00&91.00&84.00&85.00&86.00&80.00&92.00&79.00&80.00&94.00&92.00&93.00&91.00&81.00&87.00  \\
&MTCN without NTCCA  &93.68&85.64&94.31&82.49&82.00&77.58&92.47&84.09&85.84&87.13&82.71&91.80&79.03&81.00&95.00&91.49&94.68&90.73&81.06&86.84  \\
&MTCN with NTCCA &\textbf{94.21}&\textbf{86.73}&\textbf{95.67}&\textbf{83.51}&\textbf{82.43}&\textbf{78.85}&\textbf{93.68}&\textbf{84.93}&\textbf{87.00}&\textbf{88.39}&\textbf{84.11}&\textbf{92.77}&\textbf{80.00}&81.45&\textbf{95.73}&\textbf{92.38}&\textbf{95.69}&\textbf{91.75}&\textbf{82.00}&\textbf{88.04}  \\
\bottomrule[2pt]
\end{tabular}}
\label{tab:3}
\end{table*}
\subsection{Results}
 The results obtained for CelebA and LFWA by the proposed approach and several state-of-the-art approaches \cite{7410782}, \cite{rudd2016moon}, \cite{hand2017attributes}, \cite{Han2017Heterogeneous}, \cite{gunther2016affact}, and \cite{ding2017deep} are presented in Table \ref{tab:3}. The MTCN with NTCCA outperforms \cite{7410782}, \cite{rudd2016moon}, \cite{hand2017attributes}, \cite{Han2017Heterogeneous}, \cite{gunther2016affact}, and \cite{ding2017deep} for most of the 40 face attributes in both the CelebA and LFWA. For the CelebA results, in terms of the average accuracies, our MTCN with NTCCA improves on \cite{7410782} by 5.67\%, on \cite{rudd2016moon} by 2.03\%, on \cite{hand2017attributes} by 1.68\%, on \cite{Han2017Heterogeneous} by 0.37\%, on \cite{gunther2016affact} by 1.96\%, on \cite{gunther2016affact} (unaligned) by 2.65\%, and on \cite{ding2017deep} by 1.74\%. For the LFWA results, our MTCN with NTCCA improves on \cite{7410782} by 4.11\%, on \cite{hand2017attributes} by 1.65\%, and on \cite{Han2017Heterogeneous} by 1.81\%. Although our MTCN achieves  better performance among these compared methods, we do not know how much of an effect our MTCN with NTCCA has on the performance of the whole or some attribute predictions and whether our MTCN has worked on the related face attributes. Therefore, we conduct a further analysis based on Table \ref{tab:3} in the following sections.
\subsubsection{Ablation Analyses on the CelebA Dataset}
 We do not expect to see an increase in performance with MTCN for every attribute because some attributes do not have strong relationships with others, but most attributes achieved better estimations compared to the state-of-the-art methods. From the prediction presented in Table \ref{tab:3}, these attributes can be divided into three major categories based on the results of our method: \textbf{I}) attributes ($\sharp$ 1, 5, 6, 7, 10, 11, 14, 15, 16, 17, 18, 21, 23, 25, 27, 30, 31, 36, 39) that are relatively easy to predict using our MTCN; most of the results exceed 95\%, but those achieved using the compared methods are lower than 95\%. Each of these attributes is correlated with one or more other attributes, and our MTCN excavates these correlations in different levels, which is one of the most important reasons that it can obtain the best performance of all; for example, $\sharp$ 25 ($NoBeard$) relates to \{$\sharp$ 2 ($ArchedEyebrows$), 3($BushyEyebrows$), 6($Bald$), 7($Bangs$), 10($BrownHair$), 11($GrayHair$), 12($BigLips$), 18($Goatee$), 19($HeavyMakeup$), 20($HighCheekbones$), \\
 22($Male$))\}; \textbf{II}) the estimation of attributes ($\sharp$ 26 and 28) is less than 80\%; they are easily influenced by the shooting angle and pose, and few of the attributes are highly related; and \textbf{III}) these attributes are related to the attributes in \textbf{I}. Most of the time, the attributes in \textbf{III} can enhance the features of the attributes in \textbf{I}, while those of \textbf{III} benefit less from those of \textbf{I}. For example, \{$\sharp$ 2 ($ArchedEyebrows$) and $\sharp$ 25 ($NoBeard$)\}, \{$\sharp$ 3 ($BushyEyebrows$) and $\sharp$ 25 ($NoBeard$)\}, \{$\sharp$ 20 ($HighCheekbones$) and ($\sharp$ 25 ($NoBeard$), 32 ($Smiling$))\}.

\begin{table}[!hbpt]
\centering
\caption{The average accuracies of the three categories on CelebA.}
\scalebox{0.65}[0.60]{%
\begin{tabular}{c|c|c|c}
\toprule[2pt]
\textbf{Methods} & \textbf{Category I}&\textbf{Category II}&\textbf{Category III}\\\midrule[1pt]
LENet+ANet\cite{7410782}&83.42\%&77\%&85\%\\
\hline
MOON\cite{rudd2016moon}&95.46\%&76.1\%&87.99\%\\
\hline
MCNN+AUX\cite{hand2017attributes}&95.47\%&75.31\%&88.18\%\\
\hline
DMTL\cite{Han2017Heterogeneous}&95.14\%&75.56\%&87.06\%\\
\hline
AFFACT\cite{gunther2016affact}&95.63\%&76.19\%&88.41\%\\
\hline
AFFACT Unaligned\cite{gunther2016affact} (unaligned)&95.63\%&76.66\%&88.5\%\\
\hline
PaW\cite{ding2017deep}&97.31\%&78\%&89.42\%\\
\hline
MTCN without NTCCA&96.46\%&77.08\%&89.01\%\\
\hline
MTCN with NTCCA&97.58\%&78.49\%&89.88\%\\
\bottomrule[2pt]
\end{tabular}}
\label{tab:025}
\end{table}


  Table \ref{tab:025} presents the average accuracies of the methods for the three categories. For category \textbf{I}, the average accuracies of our MTCN without NTCCA are 96.46\%, which improves on \cite{7410782} by 13.04\%, on \cite{rudd2016moon} by 1\%, on \cite{hand2017attributes} by 0.83\%, on \cite{gunther2016affact} by 0.99\%, on \cite{gunther2016affact} (unaligned) by 1.32\%, and on \cite{ding2017deep} by 0.83\%. With NTCCA, MTCN improves the average accuracy by 1.12\% compared to that without NTCCA, and it shows better performance than does \cite{Han2017Heterogeneous}. Then, for category \textbf{II}, for the average accuracies of \cite{7410782}, \cite{rudd2016moon}, \cite{hand2017attributes},  \cite{Han2017Heterogeneous}, \cite{gunther2016affact}, \cite{gunther2016affact} (unaligned), \cite{ding2017deep}, our MTCN without NTCCA, and our MTCN with NTCCA  are 77\%, 76.1\%, 76.66\%, 78\%, 75.31\%, 75.56\%, 76.19\%, 77.08\%, and 78.49\%, respectively. We find that our MTCN with NTCCA achieves the best performance. Finally, for category \textbf{III}, the average accuracy of MTCN with NTCCA is 89.88\%, which improves the average accuracy by 0.99\% compared to MTCN without NTCCA, and it exceeds all of the compared methods listed above.

 Based on the analysis above, we can learn that if the face attribute relates to the others and MTCN is trained with a large enough dataset, the proposed method can show good performance via excavating the correlations among these attributes, such as the performance on categories \textbf{I} and \textbf{III}. Without NTCCA, the performance of our MTCN is nearly the same as that of the state-of-the-art methods, mostly because of the novel design of the network, which not only fully considers the differences among the face attributes but also extracts related information  to enhance itself. Further, it attempts to maximize the correlation among the high-level features through the NTCCA. Compared with the state-of-the-art methods on CelebA, our method not only improves the average accuracy of the attributes taken as a while but also greatly increases the poor accuracies of single attributes predicted by the compared methods; for example, the predictions for the attribute $Bangs$ are nearly 72\%, while that of our MTCN is 95.44\%.

\begin{table}[!hbpt]
\centering
\caption{The average accuracies of the three categories on LFWA.}
\scalebox{0.65}[0.60]{%
\begin{tabular}{c|c|c|c}
\toprule[2pt]
\textbf{Methods} & \textbf{Category I}&\textbf{Category II}&\textbf{Category III}\\\midrule[1pt]
LENet+ANet\cite{7410782}&89.17\%&74\%&79.76\%\\
\hline
MCNN+AUX\cite{hand2017attributes}&91.38\%&77.39\%&82.39\%\\
\hline
DMTL\cite{Han2017Heterogeneous}&91.67\%&75\%&81.95\%\\
\hline
MTCN without NTCCA&91.81\%&77.58\%&82.96\%\\
\hline
MTCN with NTCCA&92.77\%&78.85\%&84.26\%\\
\bottomrule[2pt]
\end{tabular}}
\label{tab:026}
\end{table}

\subsubsection{Ablation Analyses on the LFWA Dataset}
Compared with CelebA, LFWA is a relatively small dataset, so all of the average accuracies are lower than those on CelebA. Although our MTCN achieves the best performance of all of the compared algorithms, the trends in the accuracies of some of the attribute predictions are not the same as those in CelebA. For example, the accuracy of $Bangs$  ($\sharp$ 7) on LFWA is 84.51\%, and it belongs to category \textbf{II}, but $Bangs$ is in category \textbf{I} on CelebA, and $BlondHair$ ($\sharp$ 9) is in category \textbf{I} on LFWA but belongs to category \textbf{II} on CelebA. Although LFWA is a small dataset, the accuracies of most of the attributes decrease slightly compared with those on CelebA. The augmentation scheme on LFWA is an important reason, but a more important reason is attributed to the novel structure of considering the correlations of the attributes in different levels.

 Without loss of generality, we still divide these attributes into three categories according to the results of our MTCN on LFWA. Comparing the three categories with those on CelebA, we can learn that our MTCN is effective in the case of a small number of images. The details are as follows: \textbf{I}) for attributes ($\sharp$ 5, 6, 9, 10, 11, 16, 18, 19, 20, 21, 23, 27, 30, 32, 35, 36, 37, 38, 40), most of the results exceed 90\%, but those of the compared methods are lower than 90\%; \textbf{II}) the estimation of attribute ($\sharp$ 26) is less than 80\%; and \textbf{III}) for attributes ($\sharp$ 1, 2, 3, 4, 7, 8, 12, 13, 14, 15, 17, 22, 24, 25, 28, 29, 31, 33, 34, 39), all results exceed 80\%. Table \ref{tab:026} shows the detailed average accuracies of the three categories.

In terms of the attributes in category \textbf{I}, those on CelebA include attributes ($\sharp$ 1, 5, 6, 7, 10, 11, 14, 15, 16, 17, 18, 21, 23, 25, 27, 30, 31, 36, 39), while LFWA has attributes ($\sharp$ 5, 6, 9, 10, 11, 16, 18, 19, 20, 21, 23, 27, 30, 32, 35, 36, 37, 38, 40). This result indicates that attributes ($\sharp$ 1, 7, 14, 15, 17, 25, 31, 39) do not belong to category \textbf{I} from CelebA in LFWA but that attributes ($\sharp$ 9, 19, 20, 32, 35, 37, 38, 40) are in category \textbf{I}, which belongs to \textbf{III} in CelebA. The size of the dataset affects the predictions of attributes ($\sharp$ 1, 7, 14, 15, 17, 25, 27, 31, 39), but our MTCN minimizes that influence by making full use of the correlations among the attributes. Additionally, to the advantage of our system, the predictions of attributes ($\sharp$ 9, 19, 20, 32, 35, 37, 38), which are relatively difficult to predict, are not strongly affected by the size of the dataset.

In conclusion, face attributes are related, and the degrees of correlation among the different attributes are different. Excavating the related information at different levels can improve the performance of the attribute predictions. MTCN attempts to capture the correlation from different levels of features among the different attributes, such as sharing information in low-level feature layers and splitting it in the high-level feature layers while extracting related information from other subnetworks to enhance its own useful features and excavating the correlations of high-level features. These are the main reasons why our MTCN can achieve better performance on a relatively small dataset, even if it is used without NTCCA, and why the overall performance of our system on LFWA is close to that for the same attribute on CelebA. Because of its novel design compared with other methods, MTCN not only achieves the best performance, but also greatly improves the accuracies of the predictions on some single attributes, such as $Bangs$ and $Blurry$.

\section{Conclusions}

This paper presents a novel multi-task tensor correlation neural network (MTCN) for facial attribute prediction. Compared to the existing approaches, the proposed method fully explores the correlations at different levels, including sharing information in the low-level feature layers, splitting that in the high-level feature layers while extracting related information from other subnetworks to enhance its features and excavating the correlation of high-level features with NTCCA. Then, our MTCN makes final decisions for each attribute prediction. Extensive experiments demonstrate the effectiveness of our proposed system. The experimental results show that fully exploiting the correlations among the face attributes can achieve better performance, even if the training dataset is not large enough. In the future, we will improve the hybrid systems to achieve better prediction performance for the attributes in categories \textbf{II} and \textbf{III}.

\bibliographystyle{abbrv}
\bibliography{sigproc}
\end{document}